\documentclass{article}

     \usepackage[final]{neurips_2019}

\usepackage[utf8]{inputenc} 
\usepackage[T1]{fontenc}    
\usepackage{hyperref}       
\usepackage{url}            
\usepackage{booktabs}       
\usepackage{amsfonts}       
\usepackage{nicefrac}       
\usepackage{microtype}      
\usepackage{tikz}
\usepackage{caption}
\usepackage{subcaption}
\usepackage{amsmath,amssymb,amsfonts}
\title{A memory enhanced LSTM for modeling complex temporal dependencies}

\author{
Sneha Aenugu \\
University of Massachusetts Amherst\\
\texttt{saenugu@cs.umass.edu}
}

\begin{document}

\maketitle

\begin{abstract}
In this paper, we present $\Gamma$-LSTM, an enhanced long short term memory(LSTM) unit, to enable learning of hierarchical representations through multiple stages of temporal abstractions. Gamma memory, a hierarchical memory unit, forms the central memory of $\Gamma$-LSTM with gates to regulate the information flow into various levels of hierarchy, thus providing the unit with a control to pick the appropriate level of hierarchy to process the input at a given instant of time. We demonstrate better performance of $\Gamma$-LSTM model regular and stacked LSTMs in two settings (pixel-by-pixel MNIST digit classification and natural language inference) placing emphasis on the ability to generalize over long sequences. 

\end{abstract}

\section{Introduction}

Capturing complex dependencies in temporal sequences is pivotal to processing natural language. LSTMs \cite{lstm} made great strides in this area by incorporating an adaptable memory state where the contextual information regulates the storage and deletion of information from the memory state. However LSTMs still struggle to capture syntactic dependencies in complex sentences \cite{syntax} motivating the need for better architectures capable of learning complex hierarchical and temporal representations.

Several studies extended the architecture of LSTM by enhancing its memory, either by tweaking its internal memory state \cite{cheng}, \cite{moniz} or by augmenting it with an external memory with which it can interact through attentional processes \cite{memory}, \cite{mem2}. Other studies \cite{pascanu}, \cite{zhang}, \cite{zilly}, \cite{chung} focused on making the recurrent neural network deeper to capture the hierarchical structure in the temporal sequences.

In this study, we enhance the architecture of LSTM by replacing its internal memory state with a modified gamma memory. Gamma memory \cite{devries} is a short term memory model that is formulated as a cascade of first-order leaky integrators that can adapt its internal time constants to match the temporal structure of the data. The model of gamma memory is shown in Figure (1).

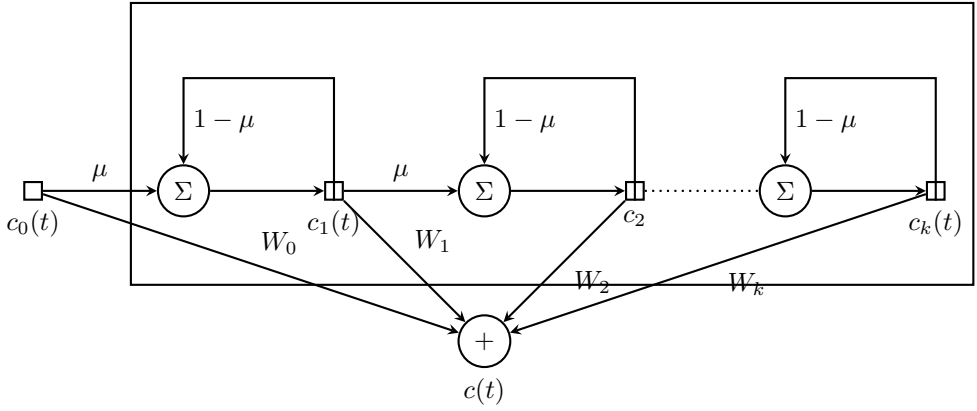
\begin{figure*}[h!]
\begin{tikzpicture}[auto, scale = 1, node distance=3cm,
  thick,main node/.style={circle,draw},
thick,small node/.style={circle,draw,scale=0.4}, rect node/.style={rectangle,draw}]

  \node[main node] (1) at (0,0) {$\Sigma$};
  \node[main node] (2) at (4,0)  {$\Sigma$};
  \node[main node] (7) at (8,0)  {$\Sigma$};
  \node[main node, label=below:$c(t)$] (8) at (4,-2)  {$+$};

  \node[rect node, label=below:$c_0(t)$] (3) at (-2,0) {};
  \node[rect node, label=below:$c_1(t)$] (4) at (2,0) {};
  \node[rect node, label=below:$c_2$] (5) at (6,0) {};
  \node[rect node, label=below:$c_k(t)$] (6) at (10,0) {};

  \draw[->,>=stealth]  (4.south) |-  (-0,1.5)  -- node {$1-\mu$}  (1.north) ;
  \draw[->,>=stealth]  (5.south) |-  (4,1.5)  -- node {$1-\mu$}  (2.north) ;
  \draw[->,>=stealth]  (6.south) |-  (8,1.5)  -- node {$1-\mu$}  (7.north) ;

  \draw[->,>=stealth]   (3) -- node {$\mu$} (1) ;
   \draw[->,>=stealth]   (1) edge  (4);
   \draw[->,>=stealth]   (4) -- node {$\mu$} (2);
   \draw[->,>=stealth]   (2) edge (5);
   \draw[->,>=stealth]   (7) edge (6);
   \draw (-0.7,-1.25) rectangle (10.5,2.5);
   \draw[dotted] (5) -- (7);
   
   \draw[->,>=stealth]   (3) -- node {$W_0$} (8) ;
   \draw[->,>=stealth]   (4) -- node {$W_1$} (8) ;
   \draw[->,>=stealth]   (5) -- node {$W_2$} (8) ;
   \draw[->,>=stealth]   (6) -- node {$W_k$} (8) ;

\end{tikzpicture}
\caption{Gamma memory - A memory network of cascaded memory cells with each memory cell gets its input from the previous cell except the first cell which gets the system input. The final memory state is a linear combination of each of the states $c_i(t)$}
\end{figure*}
The dynamics governing the information storage in the internal states of the model can be expressed by the recurrent equations \cite{gamma} shown below.
\begin{flalign}
  c_0(t) &= x(t) \\
  c_k(t) &= (1-\mu) c_k(t-1) + \mu c_{k-1}(t-1), k=1,2,...,K \\
  c(t) &= \sum_{i=0}^K W_i c_i(t)
\end{flalign}

where $x(t)$ is the input and $c_k(t)$ is the kth internal state of the memory. The final memory state $c(t)$ is represented as a linear combination of the states $c_k(t)$. The terms $\mu$ and $K$ are the parameters of the memory model corresponding to the depth and shape of the memory \cite{mozer} and $W_i$ encode the spatio-temporal correlations between the internal states $c_i$.

The above memory model is incoporated into the architecture of LSTM. In the next section, we write down the mathematical equations for the $\Gamma$-LSTM with the gamma memory as its internal state. In section 3, we apply the $\Gamma$-LSTM to two settings - digit classification and natural language inference and compare its performance against regular and stacked LSTM architectures.

\section{A memory enhanced LSTM}
The mathematical equations governing the functioning of an LSTM cell shown in Figure 2(a) are given below. 
\begin{flalign}
  i_t =& \sigma(W_{xi}x_t + W_{hi}h_{t-1} + b_i) \\ 
  f_t =& \sigma(W_{xf}x_t + W_{hf}h_{t-1} + b_f) \\
  g_t =& tanh(W_{xg}x_t + W_{hg}h_{t-1} + b_g)   \\
  o_t =& \sigma(W_{xo}x_t + W_{ho}h_{t-1} + b_o)   \\
  c_t =& f_t \odot c_{t-1} + i_t \odot g_t \\
  h_t =& o_t \odot tanh(c_t)
\end{flalign}

The internal memory state $c_t$ in Equation(8) is replaced by the equations (1) - (3). The coefficients $\mu$ for each memory cell, however, are learned by the model through non-linear gates instead of setting them to the standard values. By doing so, we allow for flexibility in the way memories in the lower level of hierarchy influence the updates in the memories higher up in the hierarchy. 

Figure 2(b) shows the LSTM cell with gamma memory as its internal state. The updated equations of the memory enhanced LSTM are given as follows.
\begin{flalign}
  i_t =& \sigma(W_{xi}x_t + W_{hi}h_{t-1} + b_i) \\
  f_{kt} =& \sigma(W_{xf}x_t + W_{hf}h_{t-1} + b_f), k=1,2,\ldots,K \\
  g_t =& tanh(W_{xg}x_t + W_{hg}h_{t-1} + b_g)   \\
  o_t =& \sigma(W_{xo}x_t + W_{ho}h_{t-1} + b_o)   \\
  c_0(t) =& i_t \odot g_t \\
  c_k(t) =& (1-f_{kt})c_k(t-1) + f_{kt} c_{k-1}(t-1), k = 1,2,\dots,K
\end{flalign}
where $i_t$ is the input gate, $f_{kt}$ is the gate modulating the memory element $c_k$, $g_t$ is the cell gate and $o_t$ is the output gate. The memory elements $c_0, c_1, \dots, c_K$ which are different levels of hierarchy are passed through a softmax attention layer to give the coefficients $a[i]$. The internal memory $c_t$ is then written as

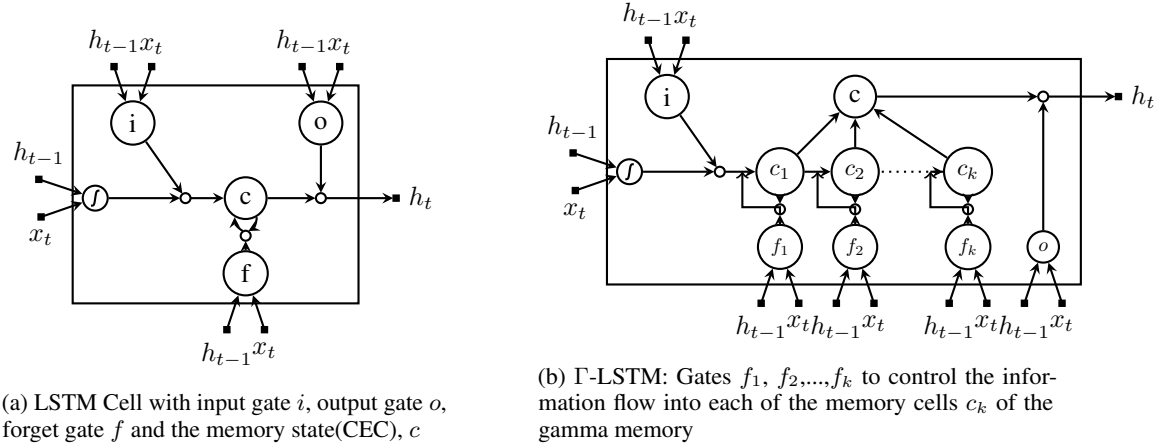
\begin{figure*}[h!]
     \centering

     \begin{subfigure}[b]{0.42\textwidth}
         \centering
        \begin{tikzpicture}[auto,scale=1,node distance=3cm,
  thick,main node/.style={circle,draw},
thick,small node/.style={circle,draw,scale=0.4}, rect node/.style={rectangle,fill,draw,scale=0.3}]

  \node[main node] (1) at (0,0) {};
  \node[small node] (7) at (1.2,0) {};
  \node[main node] (2) at (2,0)  {c};
  \node[small node] (6) at (3,0) {};
  \node[main node] (3) at (0.5,1) {i};
  \node[main node] (4) at (2,-1) {f};
  \node[main node] (5) at (3,1) {o};
  \node[small node] (8) at (2,-0.5) {};

  \node[rect node, label=above:$h_{t-1}$] (9) at (-0.75,0.25) {};
  \node[rect node, label=below:$x_t$] (10) at (-0.72,-0.25) {};

  \node[rect node, label=above:$h_{t-1}$] (11) at (0.25,1.75) {};
  \node[rect node, label=above:$x_t$] (12) at (0.75,1.75) {};

  \node[rect node, label=above:$x_t$] (13) at (3.25,1.75) {};
  \node[rect node, label=above:$h_{t-1}$] (14) at (2.75,1.75) {};

  \node[rect node, label=below:$h_{t-1}$] (15) at (1.75,-1.75) {};
  \node[rect node, label=below:$x_t$] (16) at (2.25,-1.75) {};

  \node[rect node, label=right:$h_t$] (17) at (4,0) {};

  \draw[->,>=stealth,bend left]  (8) edge (2);
  \draw[<-,>=stealth,bend right]  (8) edge (2);
  \draw[->,>=stealth]   (1) edge (7);
   \draw[->,>=stealth]   (7) edge (2);
   \draw[->,>=stealth]   (2) edge (6);
   \draw[->,>=stealth]   (5) edge (6);
   \draw[->,>=stealth]   (3) edge (7);
   \draw[->,>=stealth]   (4) edge (8);
   \draw[->,>=stealth]   (6) edge (17);
   \draw[->,>=stealth]   (9) edge (1);
   \draw[->,>=stealth]   (10) edge (1);

   \draw[->,>=stealth]   (11) edge (3);
   \draw[->,>=stealth]   (12) edge (3);

   \draw[->,>=stealth]   (13) edge (5);
   \draw[->,>=stealth]   (14) edge (5);

   \draw[->,>=stealth]   (15) edge (4);
   \draw[->,>=stealth]   (16) edge (4);
    \begin{scope}[x=0.2pt,y=0.2cm,shift={(-0cm,-0cm)}]
      \draw[] (0,0) plot[domain=-8:8] (\x,{1/(1 + exp(-\x))-0.5});
    \end{scope}

    \draw (-0.3,-1.4) rectangle (3.5,1.5);

\end{tikzpicture}

         \caption{LSTM Cell with input gate $i$, output gate $o$, forget gate $f$ and the memory state(CEC), $c$ }
     \end{subfigure}
     \hfill
     \begin{subfigure}[b]{0.49\textwidth}
         \centering
         \begin{tikzpicture}[auto, scale=1, node distance=3cm,
  thick,main node/.style={circle,draw},
thick,small node/.style={circle,draw,scale=0.4}, rect node/.style={rectangle,fill,draw,scale=0.3}]

  \node[main node] (1) at (0,0) {};
  \node[small node] (7) at (1.2,0) {};

  \node[main node,scale=0.9] (2) at (2,0)  {$c_1$};
  \node[main node,scale=0.9] (23) at (3,0)  {$c_2$};
  \node[main node,scale=0.9] (24) at (4.5,0)  {$c_k$};
  \node[main node] (28) at (3,1)  {c};

  \node[main node] (3) at (0.5,1) {i};

  \node[main node, scale=0.75] (4) at (2,-1) {$f_1$};
  \node[main node, scale=0.75] (26) at (3,-1)  {$f_2$};
  \node[main node, scale=0.75] (27) at (4.5,-1)  {$f_k$};
  \node[main node, scale=0.75] (30) at (5.5,-1)  {$o$};

  \node[small node] (8) at (2,-0.5) {};
  \node[small node] (21) at (3,-0.5) {};
  \node[small node] (22) at (4.5,-0.5) {};
  \node[small node] (29) at (5.5,1) {};

  \node[rect node, label=above:$h_{t-1}$] (9) at (-0.75,0.25) {};
  \node[rect node, label=below:$x_t$] (10) at (-0.72,-0.25) {};

  \node[rect node, label=above:$h_{t-1}$] (11) at (0.25,1.75) {};
  \node[rect node, label=above:$x_t$] (12) at (0.75,1.75) {};

  \node[rect node, label=below:$x_t$] (13) at (5.75,-1.75) {};
  \node[rect node, label=below:$h_{t-1}$] (14) at (5.25,-1.75) {};

  \node[rect node, label=below:$h_{t-1}$] (15) at (1.75,-1.75) {};
  \node[rect node, label=below:$x_t$] (16) at (2.25,-1.75) {};

  \node[rect node, label=below:$h_{t-1}$] (17) at (2.75,-1.75) {};
  \node[rect node, label=below:$x_t$] (18) at (3.25,-1.75) {};

  \node[rect node, label=below:$h_{t-1}$] (19) at (4.25,-1.75) {};
  \node[rect node, label=below:$x_t$] (20) at (4.75,-1.75) {};

  \node[rect node, label=right:$h_t$] (31) at (6.5,1) {};

  \draw[->,>=stealth]   (17) edge (26);
  \draw[->,>=stealth]   (18) edge (26);

  \draw[->,>=stealth]   (19) edge (27);
  \draw[->,>=stealth]   (20) edge (27);

  \draw[->,>=stealth]   (26) edge (21);
  \draw[->,>=stealth]   (27) edge (22);

  \draw[->,>=stealth]   (23) edge (21);
  \draw[->,>=stealth]   (24) edge (22);

  \draw[->,>=stealth]   (23) edge (28);
  \draw[->,>=stealth]   (24) edge (28);
  \draw[->,>=stealth]   (2) edge (28);

  \draw[->,>=stealth]   (13) edge (30);
  \draw[->,>=stealth]   (14) edge (30);

  \draw[->,>=stealth]   (30) edge (29);

  \draw[->,>=stealth]   (28) edge (29);

  \draw[->,>=stealth]   (29) edge (31);

  \draw[->] (22.south) |- ++(0,1mm) -| (4,0);
  \draw[->] (21.south) |- ++(0,1mm) -| (2.5,0);
  \draw[->] (8.south) |- ++(0,1mm) -| (1.5,0);

  \draw[<-,>=stealth]  (23) edge (2);
  \draw[<-,>=stealth]  (24) edge (4,0);
  \draw[dotted]  (24) -- (23);

  \draw[<-,>=stealth]  (8) edge (2);
  \draw[->,>=stealth]   (1) edge (7);
   \draw[->,>=stealth]   (7) edge (2);
   \draw[->,>=stealth]   (3) edge (7);
   \draw[->,>=stealth]   (4) edge (8);

   \draw[->,>=stealth]   (9) edge (1);
   \draw[->,>=stealth]   (10) edge (1);

   \draw[->,>=stealth]   (11) edge (3);
   \draw[->,>=stealth]   (12) edge (3);

   \draw[->,>=stealth]   (15) edge (4);
   \draw[->,>=stealth]   (16) edge (4);
    \begin{scope}[x=0.2pt,y=0.2cm,shift={(-0cm,-0cm)}]
      \draw[] (0,0) plot[domain=-8:8] (\x,{1/(1 + exp(-\x))-0.5});
    \end{scope}

   \draw (-0.3,-1.5) rectangle (6,1.5);

\end{tikzpicture}

         \caption{$\Gamma$-LSTM:  Gates $f_1$, $f_2$,...,$f_k$ to control the information flow into each of the memory cells $c_k$ of the gamma memory }
     \end{subfigure}
     
\caption{LSTM cell and an enhanced $\Gamma$-LSTM cell}
\end{figure*}
\begin{flalign}
 c_t =& \sum_{i=0}^K a[i]c_i(t-1) \\
  h_t =& o_t \odot tanh(c_t)
\end{flalign}
Here $K$ can be interpreted as \textbf{memory order} of the architecture. Increasing $K$, increases the levels of hierarchy in the memory unit.

\section{Experiments \& Results}
We investigate the effect of replacing the standard LSTM module with the $\Gamma$-LSTM module in two different settings - Pixel-by-pixel digit classification on MNIST dataset \cite{lecun} and natural language inference on SNLI dataset \cite{bowman}, placing emphasis particularly on the ability to capture long term dependencies present in the data. We demonstrate through experiments that $\Gamma$-LSTM performs better with far fewer parameters compared to stacked-LSTM units. 

\subsection{Pixel-by-pixel MNIST}
In this experiment, we consider the 28$\times$28 MNIST image as a sequence of pixels. This sequence of pixels is the input to the network and based on the configuration of the pixel sequence, the network outputs the digit (out of 10) corresponding to the image. To test the ability of the model to process longer sequences we reshape the 28$\times$28 image into 112$\times$7, such that each image is represented by 7 sequences each of length 112. The architectures of vanila LSTM, stacked LSTM (2 \& 3 layers) and $\Gamma$-LSTM are trained with the 60,000 training images. The hyperparameters for each model are tuned to give their optimal performance and the accuracy of the best model for each of the above architectures are reported on the 10,000 test images. A $\Gamma$-LSTM of memory order 3 is considered for the above experiments.

Figure 3(a) show the plots of test accuracy for each of the above architectures against the duration of training. (The plot shows training over 10 epochs). It can be seen from the figure that $\Gamma$-LSTM converges faster and to a better testing accuracy compared to the other architectures. The final test accuracies achieved by the above architectures are tabulated in Table 1. $\Gamma$-LSTM has fewer parameters compared to the stacked architectures. A comparison of the architectural complexity for all the four architectures are tabulated in Table 2.

\subsection{Natural Language Inference}
We now apply the $\Gamma$-LSTM model to a real-world application of recognizing textual entailment. The experiments are performed on the Stanford Natural Langugage Inference (SNLI) dataset. The natural language task is to the label the relation between a set of sentence pairs, a premise and a hypothesis. LSTM architecture is used as an encoder which represents the two sentences in an embedding space and a feed-forward network is used to classify the representations into one of the categories. The feed-forward network is a fully connected neural network of 4 layers with ReLU non-linearity and with dropout enabled. We test the performance of different LSTM architectures by incorporating them in the encoder module. A $\Gamma$-LSTM of memory order 3 is considered for the experiments.

Figures 3(b) shows the accuracy achieved by each of the architectures on SNLI test dataset. In this application, $\Gamma$-LSTM gives a better performance compared to the other architectures reaching to a test accuracy of 73.29 \% whereas the vanila LSTM and the stacked LSTM achieve accuracies of 72.27\% and 71.96\% respectively.

\begin{table}[h!]
\parbox{.45\linewidth}{
\centering
\begin{tabular}{ccc}
\hline
    Architecture  & Test Accuracy (\%) \\
    \hline
    LSTM (1 layer) & 93.52\\
    LSTM (2 layers) & 96.73 \\
    LSTM (3 layers) & 96.95\\
    $\Gamma$-LSTM (Memory order 3) & 97.94 \\
    \hline
\end{tabular}
\caption{Pixel-by-pixel MNIST: Test Accuracy (after 10 epochs) for all architectures for input size = 7 and sequence length = 112}
}
\hfill
\parbox{.45\linewidth}{
\centering
\begin{tabular}{ccc}
\hline
    Architecture  & Number of Parameters \\
    \hline
    LSTM (1 layer) & 71434\\
    LSTM (2 layers) & 203530\\
  LSTM (3 layers) & 335626\\
    $\Gamma$-LSTM (Memory order 3) & 123018\\
    \hline
\end{tabular}
\caption{Pixel-by-pixel MNIST: Model parameters for all architectures for input size = 7 and sequence length = 112}
}
\end{table}

\begin{figure*}[h!]
     \centering
     \begin{subfigure}[b]{0.49\textwidth}
         \centering
         \includegraphics[scale=0.4]{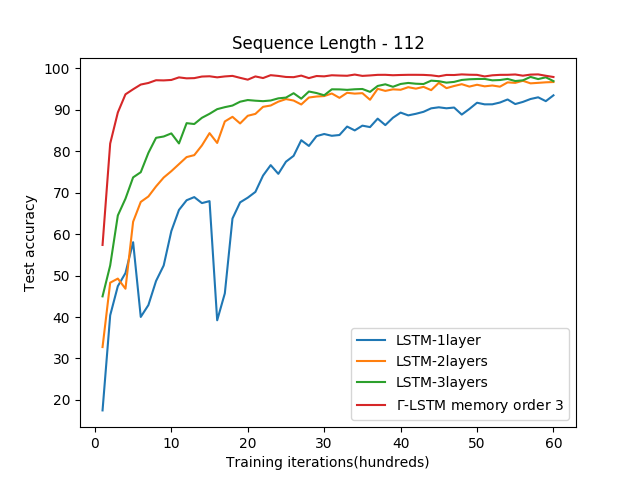}
         \caption{MNIST: Comparison of vanila LSTM, stacked LSTM (2 \& 3 layers) and $\Gamma$-LSTM plotted against training iterations for higher sequence length of 112.  $\Gamma$-LSTM achieves better performance compared to the other architectures. }
     \end{subfigure}
     \hfill
     \begin{subfigure}[b]{0.49\textwidth}
         \centering
         \includegraphics[scale=0.4]{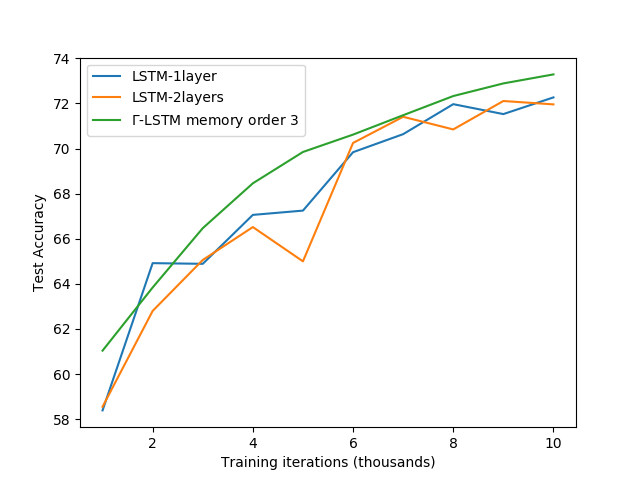}
         \caption{SNLI: Comparison of accuracy of vanila LSTM, stacked LSTM (2 \& 3 layers) and $\Gamma$-LSTM plotted against training iterations. $\Gamma$-LSTM has an overall higher test performance compared to the other architectures.}
     \end{subfigure}
     \caption{Experiments on MNIST and SNLI datasets}
\end{figure*}

\section{Conclusion}
We introduced a gamma memory unit into the LSTM architecture designing a novel recurrent memory unit, $\Gamma$-LSTM. The memory unit is hierarchical giving the model flexibility to store memories at multiple levels of hierarchy. Further, the model can choose to retrieve memories from any layer of hierarchy through an attention gating mechanism at each time step. The functioning of the memory unit is analogous to that of a band-pass filter that allows signals of certain frequencies to filter through. The model can thus allow information of higher relevance to be stored at a finer detail and store other information in a coarser format. 
 We envision further theoretical analysis on the effect of the memory unit on the recurent memory storage capacity and the incorporation of $\Gamma$-LSTM in a wide range of natural language processing and time series prediction tasks.

\bibliography{neurips_2019}
\bibliographystyle{acl_natbib}

\end{document}